Original Research Paper

# A Multi-Scale Feature Extraction and Fusion Deep Learning Method for Classification of Wheat Diseases

**[1]Sajjad Saleem, [2]Adil Hussain, [2]Nabila Majeed, [3]Zahid Akhtar and [4]Kamran Siddique**

*[1]Department of Information and Technology, Washington University of Science and Technology, USA*
*[2]Department of Physics, Kansas State University, Manhattan, KS, USA*
*[3]Department of Network and Computer Security, State University of New York Polytechnic Institute, Utica, USA*
*[4]Department of Computer Science and Engineering, University of Alaska Anchorage, Anchorage, USA*

**Abstract:** Wheat is an important source of dietary fiber and protein that is negatively impacted by a number of risks to its growth. The difficulty of identifying and classifying wheat diseases is discussed with an emphasis on wheat loose smut, leaf rust, and crown and root rot. Addressing conditions like crown and root rot, this study introduces an innovative approach that integrates multi-scale feature extraction with advanced image segmentation techniques to enhance classification accuracy. The proposed method uses neural network models Xception, Inception V3, and ResNet 50 to train on a large wheat disease classification dataset 2020 in conjunction with an ensemble of machine vision classifiers, including voting and stacking. The study shows that the suggested methodology has a superior accuracy of 99.75% in the classification of wheat diseases when compared to current state-of-the-art approaches. A deep learning ensemble model Xception showed the highest accuracy.

## Introduction

One of the main sources of calories and protein is wheat (*Triticum* spp.). One of the most significant food crops grown worldwide is wheat. The most significant grain crop is wheat, which is a staple diet for one-third of the global population. Punjab and Sindh are Pakistan's two wheat-producing regions. On the other hand, Sindh has a little greater yield per hectare than Punjab. In terms of counting, it is the top cereal crop, with rice coming in second only in terms of production and acreage. Due to numerous mechanical, physiological, and biological stresses that interfere with wheat plants' normal growth and development at all stages of development and in all natural environments, the wheat crop is vulnerable to a number of diseases that significantly lower its overall production (Krishnan *et al*., 2023; Chowdhury *et al*., 2021; Saleem *et al*., 2024). The main threats to wheat production are weather, toxicants, pollution, insects, viruses, fungi, nematodes, bacteria, and weeds. It is produced annually on an area of 219 million hectares, with an output exceeding 760 million tonnes. The USA is the world's second-largest exporter and fourth-largest producer of wheat. More than 100 distinct diseases that are brought on by different pathogens and pests target wheat. These illnesses cause about 21.5% of wheat yield to be lost each year (Singh *et al*., 2023).

Although about 200 cases of wheat diseases have been documented, the true number is unknown. More than 100 parasitic diseases that spread from plant to plant are brought on by weeds and pathogens. About 50 of them are consistently significant from an economic standpoint. All diseases are harmful overall at certain places, during certain seasons, and on certain plant sections. Diseases can affect any portion of a plant and almost any plant in any field can contract one or more diseases. All are of major concern due to their effects on the amount and/or quality of plants, straw, or grain and they attract attention due to their symptoms or indicators. Of the 50 diseases known to exist in Pakistan and have a significant economic impact, rust is the most common and harmful to wheat crops. They are found in nearly every Region that grows wheat, including every part of Pakistan. Records indicated that there had been serious outbreaks of black stem rust of wheat in Mirpur Khas, Sindh, between 1906 and 1908; yellow on stripe and orange or leaf rust of wheat in 1978 throughout Pakistan's wheat-growing regions. Significant deficiencies in crops are caused by leaf diseases spread by airborne fungi. Worldwide, wheat losses from yellow rust, leaf rust, and yellow spot can be as high as 15-40%; losses from yellow rust and stem can be as high as 40-90% (Kremneva *et al*., 2023).

One of the most dangerous biotic threats to wheat is thought to be crown and root rot, which cause enormous financial losses on a global scale. Wheat Loose Smut Complaints: On leaf blades, there are long, dark streaks that run parallel to the veins. As they age, these streaks swell, turn black, burst, and reveal black powder. The entire plant dies as a result of the impacted leaves withering, twisting, and falling. It can also occur on the sheaths, occasionally on the stem, and very infrequently on the ears. The plant often has infections in every shoot; a twisted mass of leaves replaces the ears and the plant never bears grain; if it does, the grain is generally much shriveled and useless (Srinivas *et al.*, 2024).

Symptoms of Leaf Rust, Crown, Root Rot, and Wheat Loose Smut: Foot rot disease first manifests in seedlings when brown spots appear on the lower sections of the stem and either the seeds decay in the soil or the seedlings exhibit rotting roots. The crop is thinning as a result of the damaged seedlings' eventual death. The disease known as Leaf Rust affects adult crops and is named so because it causes oval to oblong spots to grow on the lower leaves. The impacted leaves eventually turn brown as these patches get bigger. Crown and Root Rot is the term used when plants either do not produce grains or produce shriveled grains with black tips sometimes. In wheat-growing nations, a range of complex infections affecting the roots and stem base frequently reduce wheat yield, impact crop stands and degrade grain quality. Various infections, leading to damping-off, blight, tissue necrosis, and dry rot in the root, crown, sub-crown, and lower stem tissues, along with wilting and stunting in both seedlings and mature wheat plants, contribute to this complex issue. These diseases significantly affect the tiller count and the size and quantity of kernels produced within the root and stem base areas of wheat plants. Moreover, intense infections in the root and crown regions of seedlings can prove to be lethal (Bozoğlu *et al.*, 2022). In the formulation of model training, the paper contributes to the following:

- The image dataset was collected and preprocessed to convert in greyscale
- For each image, a sophisticated methodology is employed for feature calculation and fusion, encompassing binary and textural features, which includes binary and textural features, is applied to each image
- The utilization of threshold-oriented Clustering Segmentation stands out as a key technique for training Voting and Stacking classifiers on fused feature datasets
- The results were analyzed per model performance on the training and validation dataset

### Review of Literature

In this article, an automated system, the Yellow-Rust-Xception deep convolutional neural network-based model, was introduced. Diseases cause 20% of wheat production to be lost annually. Russets, smut, wheat roots, and bacterial infections are a few common wheat diseases. These diseases severely reduce plant growth and result in plant death. Early diagnosis of the disease is essential. The goal of the current study is to identify the type of yellow rust disease infection in wheat by using computer models. Based on the rust severity or percentage, the program takes an image of a wheat leaf and classifies it as resistant, susceptible, moderately sensitive, or with no disease. Convolutional Neural Networks (CNN) represent a state-of-the-art approach with layered structures inspired by the human brain. These networks can automatically learn discriminative features from data, matching or even surpassing human performance in task-specific applications. Yellow-Rust-Xception was tested, validated, and trained on a newly developed dataset of images of wheat leaves infected with yellow rust. The test accuracy was 91%. Yellow-Rust-Xception can be used to assess the severity of wheat yellow rust (Hayit *et al.*, 2021; Shafi *et al.*, 2022a).

The main goal of this study is to perform a thorough analysis of the most recent studies on Wheat Disease (WD) prediction models that have been published and discussed. The literature analysis is conducted using studies that were published between January 1997 and February 2021, adhering to Kitchenham's recommendations. After screening for inclusion/exclusion and quality rating criteria, 74 studies in total were selected. The literature lists three different types of WD: Infections caused by bacteria, fungi, or insects. The study analysis indicates that the majority of the literature's work has been found on wheat stripe rust disease (60.81%), with ANN being the most often utilized prediction technique at 13.22%. According to the findings, accuracy (67%) is the most important performance parameter and there will be the largest number of papers published on WD in 2020. Additionally, only five studies have combined SVM and NN algorithms into hybrid systems (Goyal *et al.*, 2023).

Research on four winter wheat varieties' experimental crops led to the discovery of diseases such as powdery mildew, yellow spot, septoria leaf spot, and brown rust. Except for the septoria leaf spot, we were able to track the sporulation of every disease on the list using quick spore detection equipment. The cultivar and year factors have a statistically significant independent and cumulative influence on winter wheat diseases, as demonstrated by a two-way analysis of variance. We tracked the development of wheat diseases using widely used international scales: The saari-prescot scale for yellow spot, the Peterson scale for rust infections, and the CIMMYT-developed scales for powdery mildew and septoria leaf spot. We observed sporulation of the pathogens responsible for the majority of leaf diseases, with the exception of septoria leaf spot, as a result of the 2019-2021 investigation on experimental winter wheat crops utilizing a spore-trapping apparatus. It has been established and statistically demonstrated that weather and cultivar conditions have an impact on the emergence and dissemination of fungal leaf diseases. The results show that they might be used to build a predictive model for the growth of infections in connection with the use of spore-catching devices or other comparable spore-catching equipment (Kremneva *et al.*, 2023).

The system method improves forty photos of each type of wheat leaf disease, with a focus on three classes and two types of wheat leaf diseases, such as wheat smut and wheat roots. The Image NET pretrained model is utilized for both transfer and alternating learning approaches. It also attempts to enhance the VGG16 model from the Visual Geometry Group by incorporating multi-task learning. It is demonstrated through comparison studies that this method's effects outperform those of the resnet50 model, densenet121 model, reuse-model method, and single-task model in transfer learning. According to the experimental results, the multi-task transfer learning method and enhanced VGG16 model described in this research can simultaneously identify leaf diseases in wheat, offering a dependable way to identify leaf diseases in a variety of plants. The accuracy of this model for wheat leaf diseases is 98.75% (Jiang *et al.*, 2021).

The main topic of this research is the detection of leaf diseases in wheat plants from the start of their life cycle to the finish. It emphasizes computer vision, image processing, and machine learning and shows the most effective methods for identifying different kinds of wheat leaf diseases. The primary goal is to categorize these diseases using deep CNN, a well-liked visual recognition and classification method. The article discusses several ways to categorize diseases of wheat leaves that can be seen through images, such as brown rust, stripe rust, and spot blotch. The study aims to provide an overview of the most recent methods for identifying wheat leaf diseases (Sharma and Sethi, 2023).

To provide meals for everyone, it is advisable to anticipate wheat infections early on. What diseases may be expected so early in the agricultural cycle are irreversible. The paper aims to increase farmers' awareness of the most modern methods for reducing leaf infections in wheat plants. Greater wheat yields will be produced with timely treatment and applying the proper insecticides. The farmer will have less work because early diagnosis of wheat diseases allows them to decide whether to continue producing the crop (Kumar and Kukreja, 2022).

Throughout the agricultural process, the crops require constant inspection and knowledge of the characteristics of wheat leaves. This is an expensive and time-consuming task. Thus, this research aims to identify leaf diseases, differentiate them from healthy crops without the need for constant monitoring, and decrease yield losses. One restorative class and two leaf disease classes-septoria and stripe rust-make up the three classes of the wheat leaf dataset that were employed. We presented a Support Vector Machine (SVM) model for deep learning features, which utilizes features extracted from multiple deep Transfer Learning (TL) models, applied during dataset postprocessing, and subsequently inputted into the SVM classifier. (Römer *et al.*, 2011). The components are extracted from the photos using the VGG16, VGG19, and InceptionResNetV2 models. VGG19 performs the best, achieving an accuracy of 98%, outperforming numerous studies in the literature (El-Sayed *et al.*, 2023).

Wheat diseases reduce productivity; hence, it is crucial to identify and classify them as soon as possible. This study reviewed the literature on studies published between 2017 and 2022 and provided an overview of the three primary categories of wheat diseases (bacterial, fungal, and insect-related), as well as the state-of-the-art from the previous six years of research. This research study used a variety of Machine Learning (ML) and Deep Learning (DL) algorithms to analyze 24 publications on the identification and categorization of diseases. According to the suggested research analysis, most of the datasets included in the study were self-acquired and fungal disease accounted for the majority of the attention in the literature on wheat disease. Numerous state-of-the-art models have already obtained excellent results, but many more still need to be built (Sahu and Bhat, 2023).

This wheat rust disease severely threatens food security because it can devastate the crop within a month of its first attack and reduce wheat output rates by up to 30% (Li *et al.*, 2022; Ngugi *et al.*, 2020). Because of traditional farming methods, there is a persistent worry about stopping this disease's progression as soon as possible to reduce crop losses and feed the world's expanding population. Globally, numerous precision farming options can be used to promptly detect rust attacks, lessen their disastrous consequences, and trigger site-specific corrective action. This article presents a technical summary of advanced techniques for detecting wheat rust, encompassing remote sensing, machine learning, deep learning, and the Internet of Things. The authors also examine the challenges and limitations of these methods to emphasize their practical implications (Shafi *et al.*, 2022b).

Wheat stripe rust is one of the most impactful and damaging diseases for wheat productivity. This study introduces a visual detection approach using hyperspectral imaging to enable early identification of wheat stripe rust. Over 15 days, a portable SPAD-502 chlorophyll meter recorded the chlorophyll content (SPAD value) in hyperspectral images of wheat leaves affected by stripe rust. The spectral reflectance of these samples was extracted from hyperspectral images through image segmentation based on a leaf mask. Effective wavebands were selected using principal component analysis loadings (PCAloadings) and the Successive Projections Algorithm (SPA). Then, a regression model for wheat leaf SPAD values was developed using a Backpropagation Neural Network (BPNN), with the full spectrum and selected effective wavelengths as inputs. The results indicated that the PCA-Loadings-BPNN model achieved the highest performance, with modeling and validation accuracies of 0.921-0.918, respectively, and an RPD of 3.363. This model extracted only 3.12% of the total wavelengths, simplifying the structure and greatly enhancing operational speed. Lastly, the optimal model was employed to estimate SPAD values for each pixel in the wheat leaf image, generating a spatial distribution map of chlorophyll content. This visualized map demonstrated a novel technique for early detection of wheat stripe rust, showing that infected leaves could be identified six days postinoculation and at least three days prior to visible symptoms. Yao *et al.* (2019).

## Materials and Methods

The Large Wheat Disease Classification Dataset (LWDCD2020) was utilized, comprising approximately 4,500 images that include three types of wheat diseases and one healthy category. Ozguven and Adem (2019). The dataset has four categories: Wheat loose smut, leaf rust, crown and root rot, and healthy wheat. In basic terms, each disease and healthy image in the dataset has been labeled by the original category, allowing the machine learning model to make connections between various classes of wheat conditions. In terms of preprocessing, this study initiates the data and label lists and then proceeds to iterate through all image paths. We trained the model on 80% of the LWDCD2020 dataset and tested it on the remaining 20%. The performance metrics, including accuracy, precision, recall, and F1score, were computed based on the test data to ensure that they reflect the model's generalization ability rather than its performance during training. Subsequently, we load and preprocess the image using the function where conversion to greyscale took place. The preprocessing steps involve channel swapping for OpenCV to achieve compatibility with Keras and resizing the image to 224×224 pixels.

### Proposed Methods

The proposed technique is designed to achieve improved and extracted classification of four wheat classes, including wheat load smut, leaf rust, crown and root rot, and healthy wheat. This is done through a combination of Multi-Scale Features and Image Segmentation algorithms. We have used the Canny edge detection model to detect edges in each image for selecting the Regions of Interest (RoI) (Sunil *et al.*, 2022; Tan and Le, 2019). To ensure the reliability and robustness of the proposed method, we conducted a k-fold cross-validation (with k = 5) during the training phase. This approach involved dividing the training data into five subsets, training the model on four subsets, and validating it on the remaining one. This process was repeated five times, using four subsets for training and the remaining one for validation. The process was repeated five times, with each subset serving as the validation set once and the model's average performance across all folds was calculated and reported. Additionally, future work will involve validating the model on independent datasets collected from different geographic regions and environmental conditions to further ensure its generalizability and to mitigate the risk of overfitting. A two-way threshold was defined (100, 100) to segment the object area in the image, as shown in Fig. (4). After that, we calculated binary features by selecting the area and textural features. These features include correlation, contrast, energy, and homogeneity, as shown in Fig. (1). This novel method addresses the complex problem of identifying and analyzing Wheat diseases in various environmental situations. We ensembled the three deep learning models and trained them on the calculated fused features. The overall flow of the methodology can be seen in Fig. (5).

### Multiple Feature Extraction

Our technique begins with obtaining a complete dataset, followed by severe preprocessing methods meant to maintain the highest data quality standards. Following this preliminary step, we employ sophisticated picture segmentation algorithms to precisely delineate ROIs as shown in Fig. (1).

This intelligent segmentation allows us to narrow in on the most essential locations, increasing the precision of fracture identification. Figure (4), the illustration of segmented images can be shown along with the original image.

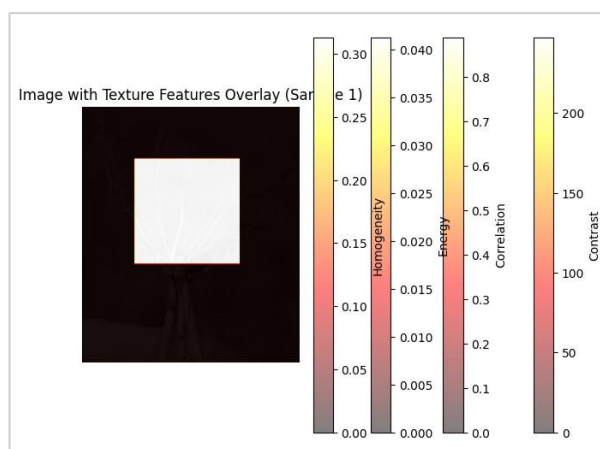

**Fig. 1:** Overlapping ROI for feature calculation

We extracted a complete collection of characteristics from each leaf picture to create a multi-feature dataset of plant photographs. These qualities can be shown in a first-order histogram of binary and textural features and the feature set explicitly includes five textural features: This category contains five unique perspective characteristics in the Gray-Level Co-occurrence Matrix (GLCM).

### Binary Feature Histogram

The binary traits, which include projection, thinness, aspect ratio, Euler number, center area, area, and axis of least second moments, are form features. For this comparison, projections with balanced width and height set at (8, 8) were used to obtain 16 binary object attributes; total calculated features for each class are as follows: Crown and root rot binary feature: 14647, leaf rust binary feature: 5480944, wheat loose smut binary feature: 27467 and healthy wheat binary feature: 6567570. Equation (1) defines the dimensions of the $i$th object (AI) as follows. This ratio tells us how much an object has stretched or contracted. An extended object such as a triangle will have an aspect ratio larger than 1, whereas the entire circle must have an overall aspect ratio of 1:

$$AspectRatio = \frac{Height}{Width} \qquad (1)$$

It is usual practice to use the following formulae to compute the centroid (($x$ $c$, $y$ $c$)) of an item concerning rows and columns:

$$xc = \frac{\sum i \sum jxi}{N} \qquad (2)$$

$$yc = \frac{\sum i \sum jyij}{N} \qquad (3)$$

The Fig. (2) histogram displays the spectrum of the values in the context of binary characteristics ($0s$ and $1s$). Typically, the values (0 and 1) are represented by the x-axis, whereas the y-axis shows the occurrence or quantity of each value.

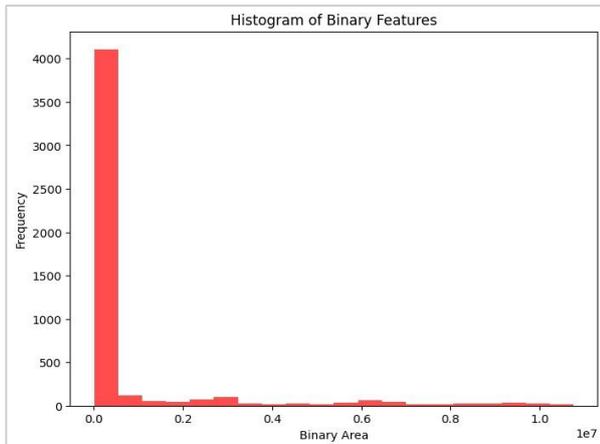

**Fig. 2:** Histogram of binary features

### Textural Feature Histogram

In this study, 5-pixel distance, entropy, inertia, correlation, inverse difference, and energy as textural features were extracted from each image using a level co-occurrence matrix. The total numbers of components are as follows: Entropy, inertia, correlation, and energy: [247.5981828909223, 0.8901628598002097, 0.041270156269148116, 0.31352326971308825], The entropy in texture features is defined using the following Eq. (4):

$$Entropy = -\sum i, jP(i,j).Log_2 P(i,j) \qquad (4)$$

The term inertia defines the contrast of an image in Eq. (5):

$$Inertia = \sum i, j(i-j)2.P(i,j) \qquad (5)$$

The correlation feature defines the value of a pixel similarity or distance in Eq. (6):

$$Correlation = \sum i, j(i-i)(j-j).\frac{P(i,j)}{\sigma i}.\sigma j(vi) \qquad (6)$$

Inversely, the homogeneity difference in an image is defined using Eq. (7):

$$InverseDifference = \sum i, j1 + \vee i - j \vee 1.P(i-i) \qquad (7)$$

However, feature extraction is a fundamental component of our novel methodology, which entails extracting binary and textural characteristics from disease-segmented areas. The disease area of an image can be visualized in Fig. (3). This thorough procedure guarantees that we collect detailed information from the targeted locations, which contributes to the efficacy and resilience of our wheat disease classification technology.

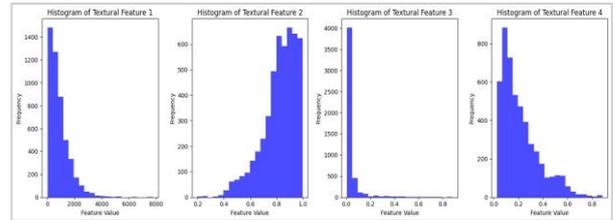

**Fig. 3:** Histogram of textural features

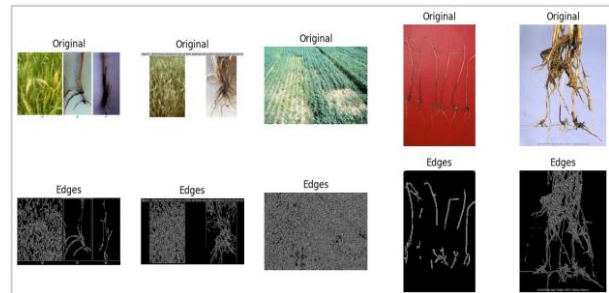

**Fig. 4:** Segmented images

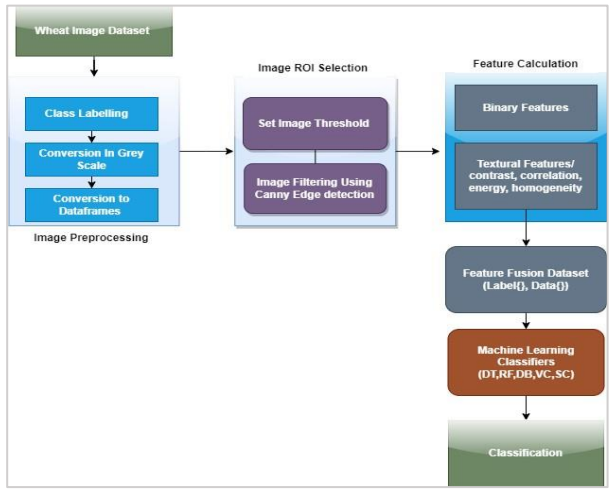

**Fig. 5:** The overall flow of the methodology

## Results and Discussion

In this study investigating the three categorizations of wheat diseases, we used Voting and Stacking ensemble machine vision classifiers posing decision trees, random forests, and Gradient boosting along with Xception, Inception V3, and ResNet 50 as deep learning models on fused multi-feature datasets.

### Voting Classifier

At first, the Voting-based classifier was employed for training and it was observed that the Voting-based classifier had the highest accuracy of 98%, precision of 98%, and F1 score of 98%. However, it shows better results than single Gradient-boosting classifiers. A confusion matrix of the Voting-based classifier shows the classification report of each class in Fig. (6).

### Stacking Classifiers

In training, the stacking-based classifiers have shown promising results, as they showed an overall accuracy of 98.1, precision of 98%, recall, and F1-score of 99%. Overall, the stacking classifiers show better results than single Gradient boosting and the highest precision of 100% than the voting classifier. The confusion matrix can be seen in Fig. (7).

### Deep Learning Modules

The transfer learning modules were trained on the intermediate layer for custom classification of Wheat disease classes. However, we have made an ensemble of three models using the loop for each model on the validation dataset. These models were trained after making a fusion of all calculated features. We defined two-dimensional arrays; array one store all the relevant features and array two stores the calculated features. The accuracy for the Xception was 99.75%, ResNet50 58.85%, and DensNet showed 98.90%.

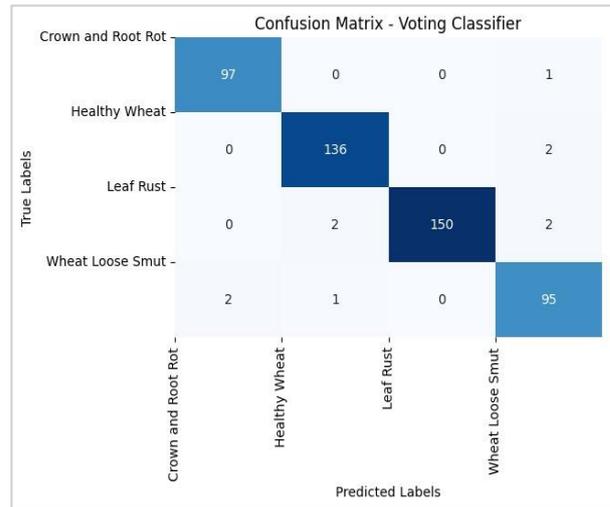

**Fig. 6:** Confusion matrix of voting-based classifier

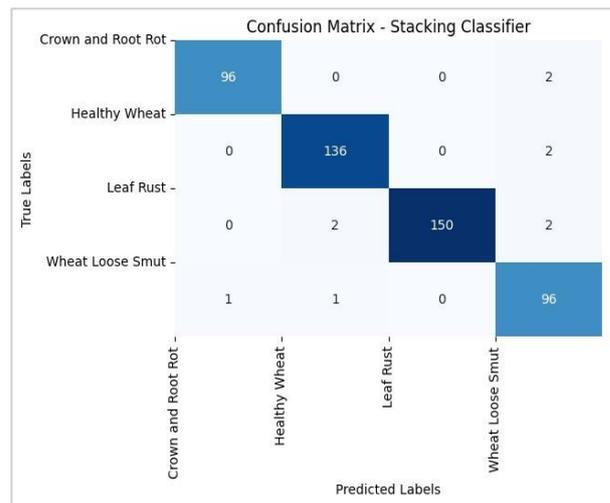

**Fig. 7:** Confusion matrix of stacking-based classifier

### Discussion

Overall, the ensemble deep learning module showed the highest accuracy in terms of training; Figs. (8-9) illustrate the overall results (Bravo *et al.*, 2003). The methodology of feature fusion contributed to obtaining the overall results. We calculated five distinct features, including binary and textural elements, such as Entropy, Inertia, Correlation, and Energy. Moreover, preprocessing included scaling to 224×224 pixels, label extraction, and grayscale conversion. The suggested technique optimized wheat class identification in various conditions by combining Multi-Scale Features with Image Segmentation techniques. In the context of the proposed methodology, the study carefully selected the Region of interest by defining the two-way thresholds in the segmentation algorithm. The suggested method demonstrated its effectiveness by utilizing deep learning models (Xception, Inception V3, and ResNet 50) in conjunction with ensemble machine vision classifiers that use decision trees, random forests, and Gradient boosting on fused multi-feature datasets. According to the results, the voting-based classifier obtained 98% accuracy, precision, and F1 score. Stacking classifiers achieved a total accuracy of 98.1-100% precision, outperforming simple Gradient Boosting. The ensemble of deep learning modules, in particular, showed the highest accuracy of 58.85% for ResNet50, 98.90% for DensNet, and 99.75% for Xception.

The ensemble's strong performance had been made possible by the thorough extracting of binary and textural aspects and feature fusion. The research offers a thorough method for correctly classifying wheat diseases by combining sophisticated deep-learning algorithms with conventional classifiers.

The comparison with the literature methodologies depicts the stand-alone performance of our feature fusion approach with an accuracy of 99.75% on a large wheat disease classification dataset, including the classes wheat loose smut, leaf rust, crown and root rot and healthy wheat (Huang *et al.*, 2020). Table (1) describes the other approaches, classification methodologies, and their model's accuracy.

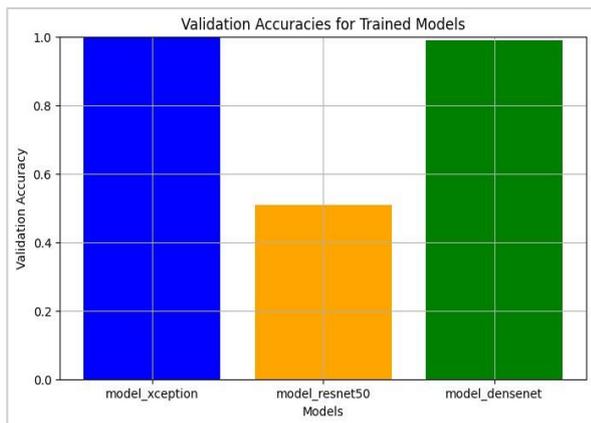

**Fig. 8:** Deep learning model comparison

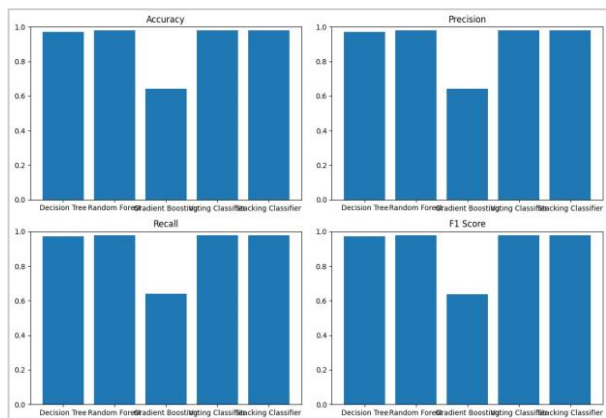

**Fig. 9:** Overall comparison of models

**Table 1:** Literature Comparison

| Ref. | Methodology | Feature extraction technique | Dataset | Performance evaluation | Overall accuracy |
|---|---|---|---|---|---|
| Hayit *et al.* (2021) | Yellow-RustXception Model | CNN | Wheat leaves infected with yellow rust | Accuracy, Precision, and Recall | 91% |
| Jiang *et al.* (2021) | Multi-Task Transfer Learning with VGG16 | PreTrained Networks | 40 photos each of wheat smut, wheat roots and more | Accuracy, Precision, F1-score, and Recall | 98.75% |
| Sharma and Sethi (2023) | Deep Learning with CNN | CNN | Plant Village dataset (tomatoes, peppers and potatoes) | Accuracy | 98.029% |
| El-Sayed *et al.* (2022) | VGG16, VGG19, InceptionResNetV2 Support Vector Machine (SVM) | SVM-DL Feature Model | Wheat leaf dataset (healthy, septoria, stripe rust) | Accuracy, Precision, F1-score, and Recall | 98% |
| Yao *et al.* (2019) | Hyperspectral Imaging for Wheat Stripe Rust Detection | PCA-loadings, SPA | Hyperspectral Images of wheat leaves infected by stripe rust | RP2, RC2, Accuracy and Validation accuracy | 92.1% and 91.8% |

## Conclusion

This study analyzes and discusses the important problem of wheat diseases, focusing on Wheat Loose Smut, Leaf Rust, and Crown and Root Rot. The paper presents a novel approach that combines ensemble machine vision classifiers (Voting, Stacking) with deep learning models (Xception, Inception V3, ResNet 50) and Image Segmentation algorithms. The findings show that the suggested methodology has a superior accuracy of 99.75% and that among deep learning models, Xception performs the best (Xu *et al*., 2017). The feature fusion techniques were applied on the Large Wheat Disease Classification Dataset, where five distinct features were calculated on each image and stored in the label and data array for its fusion.

However, in segmentation using canny edge detection algorithms, the binary and textural features were extracted using pre-defined threshold values. The novel methodology of this study contributed to advancing the feature extraction and classification of Wheat diseases using images. Furthermore, the trained model's voting-based classifier achieves 98% accuracy, precision, and F1 score; conversely, the stacking-based classifier outperforms it with 98.1% accuracy and 100% precision.

The results and discussion highlight the performance of ensemble classifiers. In particular, the deep learning module-Xception-achieves the highest results at 99.75% accuracy; by providing a strong methodology that blends cutting-edge deep learning models with conventional classifiers, the research makes a substantial contribution to wheat disease classification. The suggested method offers a useful tool for agricultural producers by addressing the difficulties in identifying and classifying wheat conditions. Potential directions for further research and development may include growing the dataset, investigating different deep learning architectures, and integrating real-time field applications to improve practicality. While the proposed method achieves high accuracy, future work should focus on expanding the dataset to include a wider variety of wheat leaf images from different geographic regions and environmental conditions. Additionally, comparing our approach with traditional disease detection methods will provide further insights into its potential advantages and limitations. Practical challenges such as cost, access, and infrastructure requirements should also be considered for real-world deployment. Finally, integrating the models with IoT devices for real-time monitoring presents an exciting avenue for future research, potentially offering a scalable and efficient solution for wheat disease management in agricultural settings.

## Acknowledgment

We sincerely thank the publisher for their support in publishing this research article. We are grateful for the resources and platform provided, which allowed us to share our findings with a broader audience. Your efforts are greatly appreciated.

## Funding Information

The authors have not received any financial support or funding to report.

## Author's Contributions

**Sajjad Saleem:** Contributed to conceptualization, methodology, software design, data curation, written-original draft preparation, interim review, and edited.

**Adil Hussain and Nabila Majeed:** Contributed to visualization and investigation.

**Kamran Siddique:** Contributed to validation.

**Zahid Akhtar:** Contributed to supervision, review, and finalization of the manuscript.

## Ethics

The authors affirm their commitment to addressing any ethical concerns that may arise after publication, ensuring transparency and taking corrective actions as needed.

## References

Bozoğlu, T., Derviş, S., Imren, M., Amer, M., Özdemir, F., Paulitz, T. C., Morgounov, A., Dababat, A. A., & Özer, G. (2022). Fungal Pathogens Associated with Crown and Root Rot of Wheat in Central, Eastern and Southeastern Kazakhstan. *Journal of Fungi*, *8*(5), 417. https://doi.org/10.3390/jof8050417

Bravo, C., Moshou, D., West, J., McCartney, A., & Ramon, H. (2003). Early Disease Detection in Wheat Fields Using Spectral Reflectance. *Biosystems Engineering*, *84*(2), 137–145.
https://doi.org/10.1016/s1537-5110(02)00269-6

Chowdhury, M. E. H., Rahman, T., Khandakar, A., Ayari, M. A., Khan, A. U., Khan, M. S., Al-Emadi, N., Reaz, M. B. I., Islam, M. T., & Ali, S. H. M. (2021). Automatic and Reliable Leaf Disease Detection Using Deep Learning Techniques. *AgriEngineering*, *3*(2), 294–312.
https://doi.org/10.3390/agriengineering3020020

El-Sayed, R., Darwish, A., & Hassanien, A. E. (2023). Wheat Leaf-Disease Detection Using Machine Learning Techniques for Sustainable Food Quality. In S. Mona (Ed.), *Artificial Intelligence: A Real Opportunity in the Food Industry* (pp. 17–28). Springer International Publishing.
https://doi.org/10.1007/978-3-031-13702-0_2

Goyal, T., Patil, A. R., Nahar, P., & Bhise, D. (2023). A Survey- Wheat Plant Diseases Recognition System Using Deep Learning Techniques. *2023 International Conference on Innovative Data Communication Technologies and Application (ICIDCA)*, 611–619.
https://doi.org/10.1109/icidca56705.2023.10099708

Hayit, T., Erbay, H., Varçın, F., Hayit, F., & Akci, N. (2021). Determination of the Severity Level of Yellow Rust Disease in Wheat by Using Convolutional Neural Networks. *Journal of Plant Pathology*, *103*(3), 923–934. https://doi.org/10.1007/s42161-021-00886-2

Huang, L., Li, T., Ding, C., Zhao, J., Zhang, D., & Yang, G. (2020). Diagnosis of the Severity of Fusarium Head Blight of Wheat Ears on the Basis of Image and Spectral Feature Fusion. *Sensors*, *20*(10), 2887. https://doi.org/10.3390/s20102887

Jiang, Z., Dong, Z., Jiang, W., & Yang, Y. (2021). Recognition of Rice Leaf Diseases and Wheat Leaf Diseases Based on Multi-Task Deep Transfer Learning. *Computers and Electronics in Agriculture*, *186*, 106184. https://doi.org/10.1016/j.compag.2021.106184

Kremneva, O., Danilov, R., Gasiyan, K., & Ponomarev, A. (2023). Spore-Trapping Device: An Efficient Tool to Manage Fungal Diseases in Winter Wheat Crops. *Plants*, *12*(2), 391. https://doi.org/10.3390/plants12020391

Krishnan, V. G., Saradhi, M. V. V., Dhanalakshmi, G., Somu, C. S., & Theresa, W. G. (2023). Design of M3FCM Based Convolutional Neural Network for Prediction of Wheat Disease. *International Journal of Intelligent Systems and Applications in Engineering*, *11*(2s), 203–210.

Kumar, D., & Kukreja, V. (2022). Deep Learning in Wheat Diseases Classification: A Systematic Review. *Multimedia Tools and Applications*, *81*(7), 10143–10187. https://doi.org/10.1007/s11042-022-12160-3

Li, Y., Qiao, T., Leng, W., Jiao, W., Luo, J., Lv, Y., Tong, Y., Mei, X., Li, H., Hu, Q., & Yao, Q. (2022). Semantic Segmentation of Wheat Stripe Rust Images Using Deep Learning. *Agronomy*, *12*(12), 2933. https://doi.org/10.3390/agronomy12122933

Ngugi, L. C., Abdelwahab, M., & Abo-Zahhad, M. (2020). Tomato Leaf Segmentation Algorithms for Mobile Phone Applications Using Deep Learning. *Computers and Electronics in Agriculture*, *178*, 105788. https://doi.org/10.1016/j.compag.2020.105788

Ozguven, M. M., & Adem, K. (2019). Automatic Detection and Classification of Leaf Spot Disease in Sugar Beet Using Deep Learning Algorithms. *Physica A: Statistical Mechanics and Its Applications*, *535*, 122537. https://doi.org/10.1016/j.physa.2019.122537

Römer, C., Bürling, K., Hunsche, M., Rumpf, T., Noga, G., & Plümer, L. (2011). Robust Fitting of Fluorescence Spectra for Pre-Symptomatic Wheat Leaf Rust Detection with Support Vector Machines. *Computers and Electronics in Agriculture*, *79*(2), 180–188. https://doi.org/10.1016/j.compag.2011.09.011

Sahu, N., & Bhat, A. (2023). A Survey: Machine Learning and Deep Learning in Wheat Disease Detection and Classification. *2023 7th International Conference on Intelligent Computing and Control Systems (ICICCS)*, 21–27. https://doi.org/10.1109/iciccs56967.2023.10142620

Saleem, S., Sharif, M. I., Sharif, M. I., Sajid, M. Z., & Marinello, F. (2024). Comparison of Deep Learning Models for Multi-Crop Leaf Disease Detection with Enhanced Vegetative Feature Isolation and Definition of a New Hybrid Architecture. *Agronomy*, *14*(10), 2230. https://doi.org/10.3390/agronomy14102230

Shafi, U., Mumtaz, R., Haq, I. U., Hafeez, M., Iqbal, N., Shaukat, A., Zaidi, S. M. H., & Mahmood, Z. (2022a). Wheat Yellow Rust Disease Infection Type Classification Using Texture Features. *Sensors*, *22*(1), 146. https://doi.org/10.3390/s22010146

Shafi, U., Mumtaz, R., Shafaq, Z., Zaidi, S. M. H., Kaifi, M. O., Mahmood, Z., & Zaidi, S. A. R. (2022b). Wheat Rust Disease Detection Techniques: A Technical Perspective. *Journal of Plant Diseases and Protection*, *129*(3), 489–504. https://doi.org/10.1007/s41348-022-00575-x

Sharma, T., & Sethi, G. K. (2023). Classification of Image-Based Wheat Leaf Diseases Using Deep Learning Approach: A Survey. *Journal of Scientific Research*, *15*(2), 421–443. https://doi.org/10.3329/jsr.v15i2.61680

Singh, J., Chhabra, B., Raza, A., Yang, S. H., & Sandhu, K. S. (2023). Important Wheat Diseases in the US and Their Management in the 21st Century. *Frontiers in Plant Science*, *13*, 1010191. https://doi.org/10.3389/fpls.2022.1010191

Srinivas, K., Singh, V. K., Srinivas, B., Sameriya, K. K., Kumar, U., Gangwar, O. P., Kumar, S., Prasad, L., & Singh, G. P. (2024). Documentation of Multi-Pathotype Durable Resistance in Exotic Wheat Genotypes to Deadly Stripe and Leaf Rust Diseases. *Cereal Research Communications*, *52*(1), 189–201. https://doi.org/10.1007/s42976-023-00364-8

Sunil, C. K., Jaidhar, C. D., & Nagamma, P. (2022). Cardamom Plant Disease Detection Approach Using EfficientNetV2. *IEEE Access*, *10*, 789–804. https://doi.org/10.1109/access.2021.3138920

Tan, M., & Le, Q. (2019). EfficientNet: Rethinking Model Scaling for Convolutional Neural Networks. *International Conference on Machine Learning*, 6105–6114

Xu, P., Wu, G., Guo, Y., chen, X., Yang, H., & Zhang, R. (2017). Automatic Wheat Leaf Rust Detection and Grading Diagnosis Via Embedded Image Processing System. *Procedia Computer Science*, *107*, 836–841. https://doi.org/10.1016/j.procs.2017.03.177

Yao, Z., Lei, Y., & He, D. (2019). Early Visual Detection of Wheat Stripe Rust Using Visible/Near-Infrared Hyperspectral Imaging. *Sensors*, *19*(4), 952. https://doi.org/10.3390/s19040952